\documentclass{article}
\PassOptionsToPackage{numbers,square,comma}{natbib}
\usepackage[preprint]{neurips_2026}
\usepackage{amsmath}
\usepackage{amssymb}
\usepackage{graphicx}
\usepackage{booktabs}
\usepackage{subcaption}
\usepackage{float}
\usepackage[utf8]{inputenc}
\usepackage[T1]{fontenc}
\usepackage{hyperref}
\usepackage{url}
\usepackage{amsfonts}
\usepackage{nicefrac}
\usepackage{microtype}
\usepackage{xcolor}
\title{Anchoring the Eigengap: \\Cross-Modal Spectral Stabilization 
for Sample-Efficient Representation Learning}

\author{
  \parbox{\linewidth}{\normalfont\centering
  \textbf{Nikhil J. Dhinagar}$^{1}$ \quad
  \textbf{Vidhi Chhatbar}$^{1}$ \quad
  \textbf{Chirag Jagad}$^{1}$ \quad
  \textbf{Pavithra Senthilkumar}$^{1}$ \\[0.5ex]
  \textbf{Sophia I. Thomopoulos}$^{1}$ \quad
  \textbf{Mahir H. Khan}$^{2}$ \quad
  \textbf{Sook-Lei Liew}$^{2}$ \\[0.5ex]
  \textbf{the ENIGMA-Stroke Recovery Working Group} \quad
  \textbf{Paul M. Thompson}$^{1}$
  } \\[3ex]
  \parbox{\linewidth}{\centering\small
  $^{1}$Imaging Genetics Center, Mark \& Mary Stevens Neuroimaging \& Informatics Institute, \\ Keck School of Medicine, University of Southern California, Los Angeles, CA, USA \\[0.25ex]
  $^{2}$Neuroscience Graduate Program, Mark \& Mary Stevens Neuroimaging \& Informatics Institute, \\ Chan Division of Occupational Science \& Occupational Therapy, \\Biomedical Engineering, University of Southern California, Los Angeles, CA, USA%
  }
}

\begin{document}

\maketitle

\begin{abstract}
\small
Deep vision models degrade sharply in low-data regimes, particularly in medical imaging where labeled samples are scarce. We show this arises not merely from overfitting but from a geometric failure: finite-sample noise corrupts the embedding covariance, collapsing the eigengap and limiting the number of recoverable signal-bearing modes.

We develop a spectral theory of finite-sample representation learning that quantifies the recoverable dimension $K(N)$, the number of eigenmodes that can be stably estimated from $N$ samples. Using perturbation theory and concentration bounds, we show that only modes with eigenvalues above the noise floor $\|\hat{\boldsymbol\Sigma} - \boldsymbol\Sigma\|_{\mathrm{op}} \sim \sqrt{D/N}$ are reliable, yielding a truncated Mahalanobis energy that governs classification performance. Under a power-law spectral model, this energy can be approximated by a truncated Riemann zeta function, linking eigenvalue decay to data efficiency and AUC.

Within this framework, multimodal learning acts as spectral stabilization: vision–language models impose low-rank constraints that suppress noise-dominated directions and preserve the eigengap, increasing $K(N)$ under data scarcity. Across MNIST and multi-disease neuroimaging, we show that multimodal training maintains more stable modes and improves class separation, even when unimodal models achieve comparable few-shot accuracy.

These results identify spectral collapse as a fundamental bottleneck in low-data learning. We use truncated Mahalanobis energy and $K(N)$ to diagnose encoder quality, and introduce zeta-based spectral filtering as a principled approach to improve data efficiency.
\end{abstract}

\section{Introduction}

Vision-language models (VLMs) trained via contrastive learning (e.g., CLIP, SigLIP) have shown strong zero-shot generalization. However, the geometric mechanisms underlying the sample efficiency gains from multimodal alignment remain unclear, particularly in data-limited domains such as medical imaging. In this work, we introduce a spectral framework that recasts representation learning as a finite-sample covariance estimation problem in the embedding space, providing a principled basis for analyzing and improving classifier data sufficiency and stability.

Our approach synthesizes classical results from the Davis–Kahan theorem, Weyl’s inequality, and concentration bounds due to Roman Vershynin into a unified theory linking encoder spectra to sample size. This yields explicit expressions for the number of eigenmodes that we can stably recover, characterizes how informative modes emerge as data increases, and explains spectral collapse when sample size is insufficient. The framework connects eigenvalue decay to a law involving the Riemann zeta function, showing why data starvation leads to instability when underlying eigenmodes cannot be stably recovered.

In this framework, multimodal alignment is a mechanism to stabilize the recoverable subspace. For example, a co-embedded text modality provides a low-dimensional anchor that constrains the overall embedding geometry, suppresses noise-dominated directions, and increases the eigengap. By classical results such as the Davis–Kahan theorem, a larger eigengap improves subspace stability, yielding more robust representations in low-data regimes.

To operationalize this theory, we propose a spectral evaluation protocol based on singular value decomposition (SVD), effective rank, and Mahalanobis energy. We use this to quantify how encoders stabilize embeddings as a function of sample size, leading to methods based on spectral filtering and mode selection to improve classifier performance in low-data regimes.

Our contributions are:

\begin{enumerate}
\item \textbf{A spectral theory of representation learning under finite samples.} We recast encoder training as a covariance estimation problem. Combining perturbation results with high-dimensional concentration bounds, we derive a law for the recoverable spectral dimension $K(N)$. This links encoder spectra, sample size, and classifier stability, identifying spectral collapse as a fundamental failure mode in low-data regimes.
\item \textbf{A geometric interpretation of multimodal learning as spectral stabilization.} We show that vision-language models act as spectral stabilizers that preserve the eigengap and prevent collapse by constraining embeddings to a low-dimensional semantic subspace, preserving rank and subspace stability.
\item \textbf{Strategies to improve data efficiency via spectral control.} Our framework yields practical strategies—including spectral filtering (e.g., zeta-filters) and mode selection—that suppress noise-dominated directions and expand the stable subspace. Across MNIST and multi-disease neuroimaging experiments, we show these principles explain and predict classifier behavior across architectures and data regimes.
\end{enumerate}

\section{Related Work}
\label{gen_inst}

Our work builds on classical results in high-dimensional probability, spectral perturbation theory, and random matrix theory. Concentration bounds for sample covariance estimation (e.g., \cite{vershynin2018high}) characterize how empirical spectra degrade when $N \ll D$, while Davis--Kahan \cite{davis1970rotation} and Weyl-type inequalities \cite{weyl1912asymptotische} quantify eigenvector and eigenvalue instability under finite-sample noise. These tools explain how informative modes are not recoverable when noise dominates the spectrum, relating to phase transitions such as the Marchenko--Pastur bulk and BBP transition \cite{baik2005phase}.

In parallel, work on the spectral structure of learned representations and neural network Jacobians includes neural collapse \cite{papyan2020prevalence} and effective rank metrics. Recent studies also link latent dimensionality to generalization \cite{elmoznino2024high}. These approaches show that learned representations concentrate data into low-dimensional subspaces, but largely treat spectra as descriptive diagnostics rather than quantities governed by sample size or linked to downstream classification performance. Information geometry and minimum description length (MDL) perspectives (\cite{amari1997statistical}, \cite{sun2019geometric}) represent model complexity through curvature, Fisher information, and coding efficiency, but operate in parameter space and do not directly address empirical encoder spectra under finite samples.

We recast representation learning as a finite-sample spectral estimation problem in embedding space, aligning with the goal of understanding data sufficiency and improving classifier stability. Our framework links encoder spectra to sample size through a recoverable spectral dimension $K(N)$, interprets the emergence and loss of modes as a spectral phase transition, and connects eigenvalue decay to a \emph{zeta law} \cite{thompson2026zeta}, named after Riemann’s zeta function (1859)\footnote{\tiny\url{https://www.claymath.org/millennium/riemann-hypothesis/}}, governing cumulative signal energy. This provides a principled account of spectral collapse as a dominant failure mode in low-data regimes and yields actionable strategies such as spectral filtering and rank preservation. In this formulation, multimodal alignment acts as spectral stabilization, preserving the eigengap and constraining embeddings to a stable subspace. As far as we know, this is the first framework linking encoder spectra, sample size, and classifier performance through a unified spectral law.

\section{Analyzing Embedding Geometry}
\label{headings}

To analyze deep learning models trained with limited data, we treat latent representations as random subspaces estimated from finite samples. Standard discriminative metrics (e.g., ROC-AUC) do not capture subspace stability and may reflect spurious directions when $N$ is small. We develop a spectral evaluation framework based on the empirical covariance of the embedding to quantify effective dimensionality, Mahalanobis-geometric class separation, and subspace stability as functions of $N$.

\subsection{Covariance Structure and Singular Value Decomposition (SVD)}

A key property of a learned representations is their covariance matrix, capturing variation across embedding dimensions. Let $\mathbf{X} \in \mathbb{R}^{N \times D}$ denote $N$ embeddings in $D$ dimensions, centered so each column has zero mean. We perform singular value decomposition (SVD) of $\mathbf{X}$, diagonalizing the empirical covariance $\hat{\Sigma} = \frac{1}{N}\mathbf{X}^\top \mathbf{X}$. The resulting eigenvectors define orthogonal modes of variation, and eigenvalues quantify their variance. In unconstrained models (e.g., vision-only CNNs), these modes capture the full spectrum of sample variance, combining meaningful signal with spurious variation. Leading components may not align with clinically relevant directions, and lower-variance modes are typically noise-dominated, especially with limited data. As $N$ decreases, the eigengap between signal- and noise-dominated eigenvalues shrinks, making the estimated subspace unstable. 

A spectral mode $k$ is \emph{reliably recoverable} only if its population eigenvalue $\lambda_k$ satisfies
\[
\lambda_k \gtrsim \|\hat{\boldsymbol\Sigma} - \boldsymbol\Sigma\|_{\mathrm{op}}.
\]
When this fails, eigenvectors mix with noise-dominated directions, leading to \emph{spectral collapse}, where we cannot identify informative modes.

\subsection{Effective Rank and Structural Dimensionality ($k(N)$)}

Although encoders produce high-dimensional embeddings, the effective capacity of the representation is typically much lower and governed by its spectral structure. A common global diagnostic is the \emph{effective rank} - the number of principal components required to explain a fixed fraction (e.g., 95\%) of the total variance. Even so, this measure is label-agnostic, so it captures both signal and noise, and can be artificially inflated as the Marchenko--Pastur noise bulk expands under limited data.

To isolate the semantically meaningful capacity of the representation, we introduce the \emph{structural dimensionality} $k(N)$, which counts the number of spectral modes that reliably capture class-relevant signal. For each mode $i$, let $\alpha_i^2$ denote the squared projection of the class difference onto the $i$-th eigenvector and $\lambda_i$ the corresponding sample eigenvalue. We define $k(N)$ as 
\[
k(N) = \sum_{i=1}^{D} \mathbf{1}\!\left(\frac{\alpha_i^2}{\lambda_i} \geq \tau\right),
\]
where $\mathbf{1}(\cdot)$ denotes the indicator function and $\tau$ is a fixed threshold (set to $0.1$ in our experiments).

This quantity serves as a data-driven estimate of the recoverable spectral dimension $K(N)$ derived in our theoretical framework, providing a practical approximation to the number of stably identifiable modes under finite samples. Modes that satisfy this condition correspond to directions in which signal dominates noise. Those that fail lie within the noise-dominated spectral bulk and do not contribute reliably to class separation.

\subsection{Mahalanobis Energy and Spectral Vulnerability}

To evaluate how effectively the representation space separates distinct clinical diagnoses, we analyze the Mahalanobis distance. Traditionally defined in the global feature space as $d_M^2 = (\mathbf{x} - \boldsymbol{\mu})^T \Sigma^{-1} (\mathbf{x} - \boldsymbol{\mu})$, this standard formulation does not make clear why we see structural collapse under data starvation. To expose this vulnerability, we project the distance into the spectral basis of the covariance matrix. Letting $\alpha_i$ represent the projection of the semantic class difference onto the $i$-th principal component and $\lambda_i$ be the corresponding sample eigenvalue (full derivation provided in Supplementary Material B2), the Mahalanobis energy decomposes as
\[
d_M^2 = \sum_{i=1}^{D} \frac{\alpha_i^2}{\lambda_i}.
\]

This formulation reveals why unconstrained models fail in low-data regimes: as sample eigenvalues in the Marchenko--Pastur noise tail shrink ($\lambda_i \to 0$), the corresponding terms grow without bound, causing the distance to be dominated by noise-dominated directions. To characterize this behavior, consider a power-law spectral model $\lambda_i \sim i^{-\beta}$. Under this assumption, the Mahalanobis energy scales as
\[
d_M^2 \sim \sum_{i} \alpha_i^2\, i^{\beta},
\]
so that modes in the tail are amplified polynomially. When $\beta \leq 1$, this summation yields a harmonic series, where cumulative contributions from many weak modes dominate the signal. This spectral accumulation is related to a truncated Riemann zeta function,
\[
\sum_{i=1}^{K(N)} i^{-\beta} \approx \zeta(\beta),
\]
linking eigenvalue decay to the number of modes we can recover \cite{thompson2026zeta}. The spectral slope $\beta$ governs data efficiency: steeper decay (larger $\beta$) concentrates energy in a small number of stable modes. Flatter spectra distribute energy across many poorly estimated directions, increasing sensitivity to finite-sample noise. By evaluating the Mahalanobis energy in this spectral form, we can quantify how well different architectures preserve the eigengap, control tail amplification, and maintain stable, data-efficient representations.

\subsection{Subspace Stability and the Davis-Kahan Theorem}
A failure point of representation learning in medical imaging is insufficient data (low $N$). We measure this instability using the Davis--Kahan theorem, which bounds the deviation between the population eigenvectors $\mathbf{v}_k$ of the true covariance $\Sigma$ and the empirical eigenvectors $\hat{\mathbf{v}}_k$ of $\hat{\Sigma}$. The subspace error, represented by the angle between the subspaces, is bounded by the eigengap $\delta_k$:
\[
\sin \Theta(\mathbf{v}_k, \hat{\mathbf{v}}_k) \le \frac{2 \|\hat{\boldsymbol\Sigma} - \boldsymbol\Sigma\|_{\mathrm{op}}}{\delta_k}.
\]

\begin{center}
\fbox{
\begin{minipage}{0.92\linewidth}
\small
\textbf{Zeta law (finite-sample spectral form).}
\quad
Let
\[
d_M^2 := (\mathbf{x}-\boldsymbol{\mu})^\top \Sigma^{-1} (\mathbf{x}-\boldsymbol{\mu}) = \sum_i \frac{\alpha_i^2}{\lambda_i},
\]
with $\lambda_i \sim i^{-\beta}$, and finite-sample perturbation $\|\hat{\boldsymbol\Sigma}-\boldsymbol\Sigma\|_{\mathrm{op}} \sim \sqrt{D/N}$ inducing subspace error $\sin\Theta \lesssim \|\hat{\boldsymbol\Sigma}-\boldsymbol\Sigma\|_{\mathrm{op}}/\delta_k$. Then only modes satisfying $\lambda_k \gtrsim \|\hat{\boldsymbol\Sigma}-\boldsymbol\Sigma\|_{\mathrm{op}}$ are recoverable, defining $K(N)$ and yielding
\[
d_M^2(N) \approx \sum_{i=1}^{K(N)} \frac{\alpha_i^2}{\lambda_i}
\;\sim\;
\sum_{i=1}^{K(N)} i^{-\beta}
\;\to\;
\zeta(\beta) \quad \text{as } K(N)\to\infty,
\]
where $\zeta$ denotes the Riemann zeta function.
\end{minipage}
}
\end{center}

\section{Methods}
\label{others}

To study the behavior of representation learning under finite samples, we consider two complementary settings. First, a controlled synthetic setting (handwritten digit classification) that isolates the spectral mechanisms underlying dimensional collapse. This simpler example illustrates how the embedding covariance and its spectrum evolve as a function of sample size $N$. We then extend the analysis to high-dimensional clinical neuroimaging data, where we evaluate the same spectral quantities across varying data regimes. We can then assess how different encoder architectures behave. Their performance depends on the stability of the underlying embedding subspace.

\subsection{Synthetic Proxy: Unconstrained Variance MNIST}
\begin{minipage}{0.65\linewidth}
To study representation learning in a controlled high-variance, low-signal setting, we construct an ``unconstrained variance'' variant of the MNIST dataset. Standard MNIST is dominated by structured signal (digit topology), so we introduce spurious variation by assigning a high-intensity, uniformly random RGB background to each image. Here, irrelevant variance dominates the embedding unless appropriately controlled.
\end{minipage}
\begin{minipage}{0.33\linewidth}
    \centering
    \includegraphics[width=\linewidth]{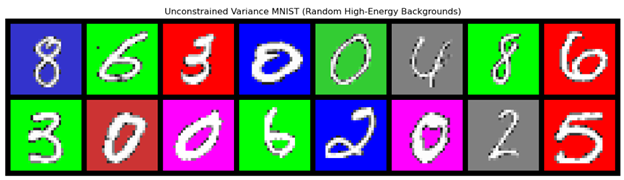}
    \captionof{figure}{Corrupted MNIST.}
\end{minipage}

We evaluate a unimodal vision encoder and a multimodal vision--language model across a data-rich regime ($N = 100\%$, 60{,}000 images) and limited data regimes ($N = 0.5\%, 1\%, 2\%$). For each, we extract test-set embeddings and compute spectral diagnostics including effective rank, Mahalanobis energy, and Davis--Kahan subspace perturbation (via $\sin \Theta$). Subspace stability is assessed by measuring the principal angle between embeddings learned on a data-rich reference sample versus those learned under reduced $N$, treating the former as a finite-sample proxy for the stable subspace.

These measures reveal a consistent pattern. In the unimodal model, spurious high-variance directions dominate the spectrum at lower training $N$, leading to large subspace deviations. Alignment with the data-rich solution is lost. In contrast, the multimodal model preserves the leading modes and maintains a stable embedding subspace across data regimes. This stabilization arises because the text encoder imposes a low-dimensional alignment constraint on the embedding: the contrastive objective concentrates variance along directions consistent with the shared image–text representation, suppressing unconstrained visual variability. \emph{The spectrum decays more rapidly, the eigengap is preserved, and a larger fraction of leading modes remain above the finite-sample noise floor}, improving subspace stability.

\subsection{Real-World Application: Multi-Disease Medical Imaging}

We extend the spectral analysis to a clinical neuroimaging setting, using 3D MRI scans labeled by diagnostic class (Alzheimer’s disease, stroke, and controls). We evaluate subspace stability and spectral structure of unimodal vision encoders and multimodal text-constrained models at varying sample sizes ($N = 5\%$ to $100\%$). Following standard protocols \cite{dhinagar2022evaluation, Tustison2010, Zhu2021}, MRIs were skull-stripped, bias-corrected, and registered to a 2-mm isotropic template (128x256x256 voxels). Table \ref{tab:dataset_summary} summarizes the datasets used in this work. To perform Alzheimer’s Disease (AD) classification, we utilized the publicly available ADNI (Alzheimer’s Disease Neuroimaging Initiative) \cite{Weiner2015} and OASIS (Open Access Series of Imaging Studies) \cite{LaMontagne2019} MRI datasets. The ADNI cohort was partitioned into 2,577 training, 302 validation, and 1,219 testing T1-weighted scans. To prevent data leakage within this longitudinal dataset, unique subjects were strictly restricted to specific folds. For stroke classification, we analyzed data from the Enhancing Neuroimaging Genetics through Meta-Analysis (ENIGMA) Stroke Recovery working group \cite{Liew2022}, one of the largest multisite retrospective stroke datasets, from 10 countries. We split this data into 1,077 training, 120 validation, and 300 test scans. 683 scans (600 from OASIS, 83 from ENIGMA-Stroke) were reserved for zero-shot testing.

\begin{table}[h!] 
    \centering
    \caption{Summary of clinical neuroimaging datasets.}
    \label{tab:dataset_summary}
    \small 
    \setlength{\tabcolsep}{4pt} 
    \begin{tabular}{l c c c c} 
        \toprule
        \textbf{Dataset} & \textbf{N Scans (Subj)} & \textbf{Age (yrs)} & \textbf{Sex (F/M)} & \textbf{N Sites} \\
        \midrule
        ADNI & 4,098 (1,188) & 55.2--95.8 & 556/632 & 62 (U.S.) \\
        OASIS & 600 (600) & 43.5--97.0 & 341/259 & 1 (U.S.) \\
        ENIGMA-Stroke & 1,580 (1,580) & 17.0--93.0 & 616/963$^*$ & 50 (Global) \\
        \bottomrule
        \multicolumn{5}{l}{\scriptsize $^*$Sex data missing for one ENIGMA-Stroke subject.}
    \end{tabular}
\end{table}

\subsubsection{Unimodal Vision-Only Baselines (Unconstrained)}
We established a baseline using architectures that lack a semantic text bottleneck trained on the T1-w MRI images: a 11M 3D DenseNet121 CNN (fully convolutional) \cite{Huang2017}, and 18M DCFormer (hybrid) \cite{xin2025med3dvlm} trained purely on binary image-to-label cross-entropy.

\subsubsection{Multimodal Vision-Language Model (Semantically Anchored)}
Our primary experimental model is a multimodal SigLIP-based architecture \cite{Zhai2023} \cite{xin2025med3dvlm} \cite{Radford2021} trained through pairwise sigmoid contrastive loss. We refer to this model class as vision-language models (VLM). 

We used 3D volumetric brain MRIs as visual inputs and diagnostic captions (e.g., ``75-year-old male diagnosed with stroke'') as the semantic anchor, employing systematic text augmentation \cite{dhinagar2025leveraging} during training to generate diverse caption variations that enrich the semantic distribution and prevent overfitting.

\textbf{Low-Dimensional Structure of Text:}
The multimodal architecture may be more data-efficient as it relies on the relatively low-dimensional structure of the text embedding space. While MRI volumes are high-dimensional and contain substantial nuisance variation, the associated text embeddings are constrained to a limited set of semantic attributes (e.g., age, sex, diagnosis). By aligning visual features, the model restricts their embeddings to a lower-dimensional, semantically coherent subspace. From a spectral perspective, this constraint reweights the covariance spectrum, concentrating variance into a smaller number of signal-aligned modes, suppressing unstructured directions. This steepens the spectral decay, preserves the eigengap, and increases the number of modes that satisfy $\lambda_k \gtrsim \|\hat{\boldsymbol{\Sigma}}-\boldsymbol{\Sigma}\|_{\mathrm{op}}$, increasing the recoverable dimension $K(N)$. This lessens the influence of noise-dominated modes in the Mahalanobis energy and improves subspace stability under finite-sample perturbations, consistent with the zeta-law model of data efficiency.

\textbf{Initialization and Training Dynamics:}
To isolate the regularizing effect of the text constraint, the 3D vision encoder in the multimodal architecture was trained from scratch. Unlike approaches that rely on unimodal pre-training to impose structural priors, the model is trained solely via the sigmoid contrastive objective over 20 epochs. This setting tests whether the low-dimensional text bottleneck can induce a well-separated spectrum—i.e., can it preserve the eigengap and stabilize the leading modes—without relying on prior structural knowledge.
    
\subsubsection{Ablation Studies}
\textbf{Architectural Invariance (Vision Encoders):} We swapped the vision backbone inside the VLM framework, tracking CNN, and DCFormer variants. This isolated the effect of the sigmoid loss, to show that subspace stability is a property of the multimodal text constraint, not a byproduct of a specific vision architecture. 

\textbf{Effect of Tokens and Missing Modalities:} Real-world clinical data frequently suffers from missing information. We ablated the robustness of the semantic anchor by introducing token masking during evaluation by
randomly masking specific demographic identifiers (age/sex). (see Appendix \ref{sec:appendix_masking})

\subsection{Discriminative and Qualitative Evaluation}

We complement spectral analysis with standard discriminative metrics and geometric visualizations to relate representation structure to downstream performance.

\textbf{Discriminative Performance (ROC-AUC):} Performance was evaluated using the Area Under the Receiver Operating Characteristic Curve (ROC-AUC). For the multi-disease case, we used a one-vs-rest (OvR) formulation and report macro-averaged $N$-way ROC-AUC alongside class-specific values. We could then assess how lowering the sample size $N$ affects diagnostic accuracy across classes.

\textbf{Subspace Visualization (UMAP):} We visualize embedding geometry by projecting latent representations into 2D and 3D using Uniform Manifold Approximation and Projection (UMAP), to visualize class structure and subspace separation.

\section{Experimental Results}
\subsection{Synthetic Proxy: Unconstrained Variance MNIST}

We first validate the spectral framework on the Unconstrained Variance MNIST dataset, which provides a controlled setting to probe spectral collapse. As shown in Table \ref{tab:mnist_data_regimes}, under severe data limitation ($N \le 2\%$), the unimodal CNN exhibits substantial instability in its structural dimensionality, with $k(N)$ fluctuating markedly ($\pm 2.10$ at $N=0.5\%$). This variability reflects the breakdown of the recoverability condition $\lambda_k \gtrsim \|\hat{\boldsymbol{\Sigma}}-\boldsymbol{\Sigma}\|_{\mathrm{op}}$, as noise-dominated modes enter the spectrum and destabilize the estimated subspace.

In contrast, the multimodal model maintains a stable recoverable dimension, with $k(N) = 9.00 \pm 0.00$ across all data regimes. This indicates that the leading eigenmodes remain consistently above the finite-sample noise floor, preserving the effective eigengap and preventing spectral collapse.

These effects are observable in the spectral diagnostics. The log--log spectral slope (Figure~3) shows that, as $N$ decreases, the unimodal spectrum flattens and the variance tail becomes dominated by noise, consistent with Marchenko--Pastur behavior. By contrast, the multimodal spectrum retains a steeper decay, concentrating variance in a smaller number of stable modes. Correspondingly, the Davis--Kahan stability curves (Figure~2) show much lower subspace perturbation, implying that the leading eigenspace remains aligned with the data-rich reference. Together, these results confirm the predicted mechanism: multimodal alignment preserves the eigengap, stabilizes the recoverable subspace, and sustains $K(N)$ under data scarcity.%

\begin{table}[t] 
\centering
\small
\begin{minipage}{0.58\linewidth}
    \centering
    \caption{\textbf{MNIST Spectral Dynamics.} Comparison of text-anchored VLM vs. unconstrained CNN (10 runs).}
    \label{tab:mnist_data_regimes}
    \setlength{\tabcolsep}{2pt} 
    \begin{tabular}{@{}l cccc@{}}
    \toprule
    \textbf{Metric} & \textbf{0.5\%} & \textbf{1\%} & \textbf{2\%} & \textbf{100\%} \\
    \midrule
    \multicolumn{5}{c}{\textit{Architecture: VLM (CNN)}} \\
    Eff. Rank & 8.200$_{\pm0.600}$ & 8.200$_{\pm0.400}$ & 8.000$_{\pm0.000}$ & 8.000$_{\pm0.000}$ \\
    $k(N)$    & 9.000$_{\pm0.000}$ & 9.000$_{\pm0.000}$ & 9.000$_{\pm0.000}$ & 9.000$_{\pm0.000}$ \\
    M-E  & 14.650$_{\pm0.390}$ & 16.760$_{\pm0.140}$ & 18.100$_{\pm0.040}$ & 20.060$_{\pm0.040}$ \\
    Test AUC  & 0.956$_{\pm0.010}$ & 0.982$_{\pm0.002}$ & 0.991$_{\pm0.001}$  & 0.998$_{\pm0.000}$ \\
    \midrule
    \multicolumn{5}{c}{\textit{Architecture: CNN (Vision only)}} \\
    Eff. Rank & 6.900$_{\pm0.300}$ & 6.900$_{\pm0.540}$ & 7.000$_{\pm0.000}$ & 9.000$_{\pm0.000}$ \\
    $k(N)$    & 8.300$_{\pm2.100}$ & 8.200$_{\pm1.830}$ & 8.800$_{\pm1.470}$ & 9.000$_{\pm0.000}$ \\
    M-E  & 18.440$_{\pm0.250}$ & 19.160$_{\pm0.150}$ & 19.720$_{\pm0.160}$ & 22.100$_{\pm0.160}$ \\
    Test AUC  & 0.991$_{\pm0.003}$ & 0.997$_{\pm0.001}$ & 0.998$_{\pm0.000}$ & 1.000$_{\pm0.000}$ \\
    \bottomrule
    \end{tabular}
\end{minipage}
\hfill
\begin{minipage}{0.38\linewidth}
    \centering
    \includegraphics[width=\linewidth]{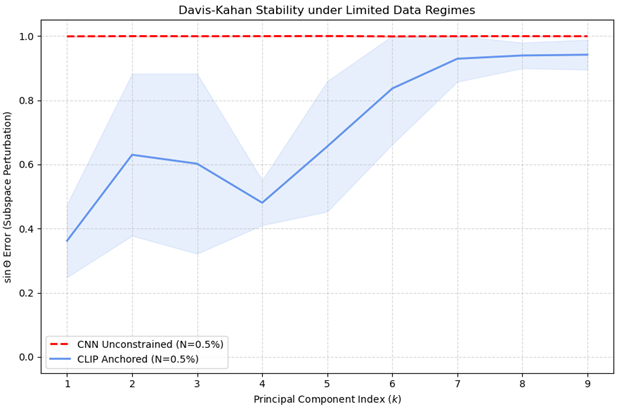}
    \captionof{figure}{\textbf{Davis--Kahan Stability.} Rotation ($\sin\Theta$) as data is reduced; lower is more stable.}
    \label{fig:davis_kahan}
\end{minipage}
\end{table}

\begin{figure}[ht!] 
    \centering
    \includegraphics[width=0.75\textwidth]{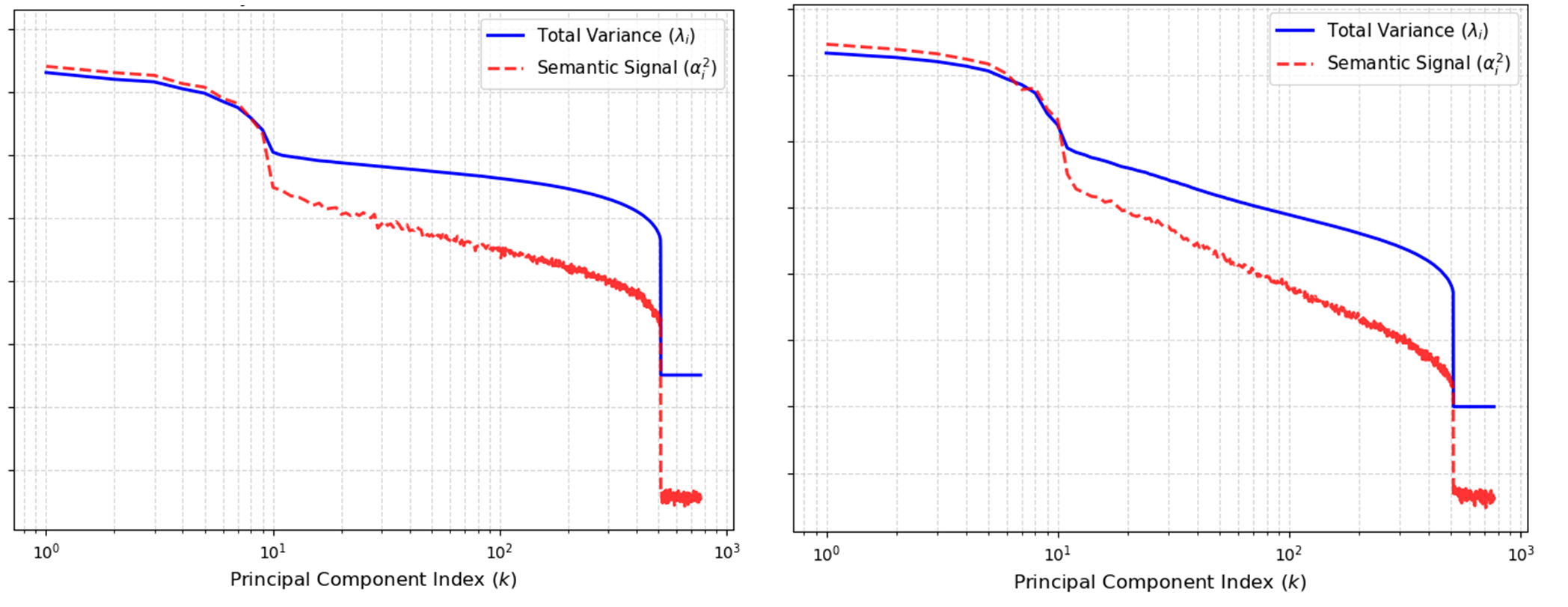}
 
    \caption{\textbf{Spectral decomposition (log--log) of the embedding covariance for the Unconstrained Variance MNIST dataset under severe data limitation ($N=0.5\%$).} VLM (left) and Vision only (right). Eigenvalues $\lambda_i$ (total variance) and $\alpha_i^2$ (signal) are shown across principal components. While the overall spectral decay is similar across models, reflecting the low intrinsic dimensionality of MNIST, the key distinction lies in signal alignment: the multimodal model maintains a higher fraction of modes with $\alpha_i^2 / \lambda_i$ above the noise floor, consistent with a larger effective $K(N)$.}
\end{figure}

\subsection{Real-World Application: Multi-disease Medical Imaging}
We extend the spectral framework to a multi-disease neuroimaging setting, tracking the evolution of the embedding covariance as the training sample size decreases from $100\%$ to $5\%$. As shown in Tables \ref{tab:dcformer_starvation_results} and \ref{tab:clip_results_dense}, standard discriminative performance (ROC-AUC) degrades with decreasing $N$, consistent with reduced data support. The CLIP model demonstrated excellent diagnostic capabilities, achieving mean ROC-AUC of 0.952 for Stroke, 0.924 for Alzheimer's, and 0.875 for Control classifications. The spectral analysis provides a complementary view, linking this degradation to changes in the recoverable subspace and the stability of the leading eigenmodes.

A key observation is the breakdown of label-agnostic dimensionality measures under data limitation. In Table \ref{tab:dcformer_starvation_results}, the Effective Rank for the cross-modal VLM + DCFormer model increases from 3 (at $100\%$ data) to 122 (at $5\%$ data). This apparent increase does not reflect greater semantic capacity; rather, it arises from the expansion of the Marchenko--Pastur noise bulk, which inflates the cumulative variance captured by weak, noise-dominated modes. As a result, variance-based thresholds (e.g., 95\%) increasingly select unstable directions that do not satisfy the recoverability condition $\lambda_k \gtrsim \|\hat{\boldsymbol{\Sigma}} - \boldsymbol{\Sigma}\|_{\mathrm{op}}$. This behavior is consistent with the spectral law: as $N$ decreases, the noise floor rises, reducing the number of modes that remain above it and decreasing the effective $K(N)$. Consequently, while global variance measures suggest increasing dimensionality, the number of stable, signal-bearing modes contracts, leading to reduced subspace stability and degraded classification performance.

\begin{table*}[t] 
\centering
\small
\setlength{\tabcolsep}{2pt}
\caption{\textbf{Spectral Dynamics and Predictive Performance in Limited Data Regime (Neuroimaging)}. Comparison of text-anchored VLM vs. unconstrained DCFormer.}
\label{tab:dcformer_starvation_results}
\begin{tabular}{@{}l ccccccc@{}}
\toprule
\textbf{Metric} & \textbf{5\%} & \textbf{10\%} & \textbf{25\%} & \textbf{35\%} & \textbf{50\%} & \textbf{75\%} & \textbf{100\%} \\
\midrule
\multicolumn{8}{c}{\textbf{Architecture: VLM with DCFormer as vision encoder}} \\
\midrule
Eff. Rank & 122.0$_{\pm5.000}$ & 120.5$_{\pm1.500}$ & 63.5$_{\pm61.500}$ & 63.5$_{\pm62.500}$ & 2.000$_{\pm0.000}$ & 2.500$_{\pm0.500}$ & 3.000$_{\pm0.000}$ \\
$k(N)$ (Stroke) & 5.000$_{\pm0.000}$ & 5.000$_{\pm1.000}$ & 3.000$_{\pm0.000}$ & 2.500$_{\pm0.500}$ & 2.500$_{\pm0.500}$ & 3.500$_{\pm0.500}$ & 4.000$_{\pm0.000}$ \\
$k(N)$ (AD) & 1.000$_{\pm0.000}$ & 2.000$_{\pm1.000}$ & 1.000$_{\pm0.000}$ & 2.500$_{\pm1.500}$ & 1.000$_{\pm0.000}$ & 1.500$_{\pm0.500}$ & 3.000$_{\pm0.000}$ \\
$k(N)$ (CN) & 0.000$_{\pm0.000}$ & 0.000$_{\pm0.000}$ & 0.000$_{\pm0.000}$ & 0.000$_{\pm0.000}$ & 0.000$_{\pm0.000}$ & 0.000$_{\pm0.000}$ & 2.000$_{\pm2.000}$ \\
M-E (Stroke) & 4.120$_{\pm0.070}$ & 4.220$_{\pm0.190}$ & 4.670$_{\pm0.580}$ & 4.470$_{\pm0.150}$ & 5.370$_{\pm0.110}$ & 5.770$_{\pm0.010}$ & 5.870$_{\pm0.060}$ \\
M-E (AD) & 2.890$_{\pm0.030}$ & 2.860$_{\pm0.150}$ & 3.190$_{\pm0.300}$ & 3.150$_{\pm0.180}$ & 3.730$_{\pm0.130}$ & 3.920$_{\pm0.060}$ & 4.000$_{\pm0.020}$ \\
M-E (CN) & 2.280$_{\pm0.070}$ & 2.310$_{\pm0.110}$ & 2.520$_{\pm0.220}$ & 2.460$_{\pm0.120}$ & 2.930$_{\pm0.140}$ & 3.170$_{\pm0.040}$ & 3.220$_{\pm0.010}$ \\
Test AUC & 0.521$_{\pm0.002}$ & 0.537$_{\pm0.019}$ & 0.651$_{\pm0.118}$ & 0.632$_{\pm0.000}$ & 0.829$_{\pm0.034}$ & 0.899$_{\pm0.024}$ & 0.917$_{\pm0.010}$ \\
OOD Test AUC & 0.437$_{\pm0.042}$ & 0.377$_{\pm0.005}$ & 0.642$_{\pm0.009}$ & 0.552$_{\pm0.083}$ & 0.706$_{\pm0.027}$ & 0.757$_{\pm0.010}$ & 0.736$_{\pm0.046}$ \\
\midrule
\multicolumn{8}{c}{\textbf{Architecture: DCFormer (vision only)}} \\
\midrule
Eff. Rank & 1.000$_{\pm0.000}$ & 1.000$_{\pm0.000}$ & 1.500$_{\pm0.500}$ & 2.000$_{\pm0.000}$ & 2.000$_{\pm0.000}$ & 2.000$_{\pm0.000}$ & 2.000$_{\pm0.000}$ \\
$k(N)$ (Stroke) & 2.500$_{\pm0.500}$ & 2.500$_{\pm0.500}$ & 2.500$_{\pm0.500}$ & 2.000$_{\pm0.000}$ & 3.000$_{\pm0.000}$ & 0.500$_{\pm0.500}$ & 1.000$_{\pm1.000}$ \\
$k(N)$ (AD) & 1.500$_{\pm0.500}$ & 1.000$_{\pm0.000}$ & 1.000$_{\pm0.000}$ & 2.000$_{\pm0.000}$ & 2.000$_{\pm0.000}$ & 2.000$_{\pm0.000}$ & 2.500$_{\pm0.500}$ \\
$k(N)$ (CN) & 0.000$_{\pm0.000}$ & 0.000$_{\pm0.000}$ & 0.000$_{\pm0.000}$ & 0.000$_{\pm0.000}$ & 2.000$_{\pm0.000}$ & 2.500$_{\pm0.500}$ & 1.500$_{\pm1.500}$ \\
M-E (Stroke) & 5.840$_{\pm0.040}$ & 5.980$_{\pm0.040}$ & 6.130$_{\pm0.030}$ & 6.320$_{\pm0.070}$ & 6.260$_{\pm0.060}$ & 6.210$_{\pm0.010}$ & 6.280$_{\pm0.000}$ \\
M-E (AD) & 4.200$_{\pm0.060}$ & 4.240$_{\pm0.010}$ & 4.300$_{\pm0.000}$ & 4.400$_{\pm0.020}$ & 4.410$_{\pm0.020}$ & 4.410$_{\pm0.030}$ & 4.420$_{\pm0.020}$ \\
M-E (CN) & 3.390$_{\pm0.020}$ & 3.480$_{\pm0.050}$ & 3.550$_{\pm0.030}$ & 3.690$_{\pm0.040}$ & 3.670$_{\pm0.010}$ & 3.660$_{\pm0.000}$ & 3.700$_{\pm0.030}$ \\
Test AUC & 0.865$_{\pm0.006}$ & 0.893$_{\pm0.003}$ & 0.916$_{\pm0.001}$ & 0.942$_{\pm0.005}$ & 0.944$_{\pm0.005}$ & 0.951$_{\pm0.001}$ & 0.947$_{\pm0.001}$ \\
OOD Test AUC & 0.696$_{\pm0.000}$ & 0.688$_{\pm0.022}$ & 0.749$_{\pm0.002}$ & 0.771$_{\pm0.002}$ & 0.786$_{\pm0.008}$ & 0.798$_{\pm0.004}$ & 0.805$_{\pm0.001}$ \\
\bottomrule
\end{tabular}
\end{table*}

\begin{figure}[hb] 
    \centering
    \includegraphics[width=0.75\textwidth]{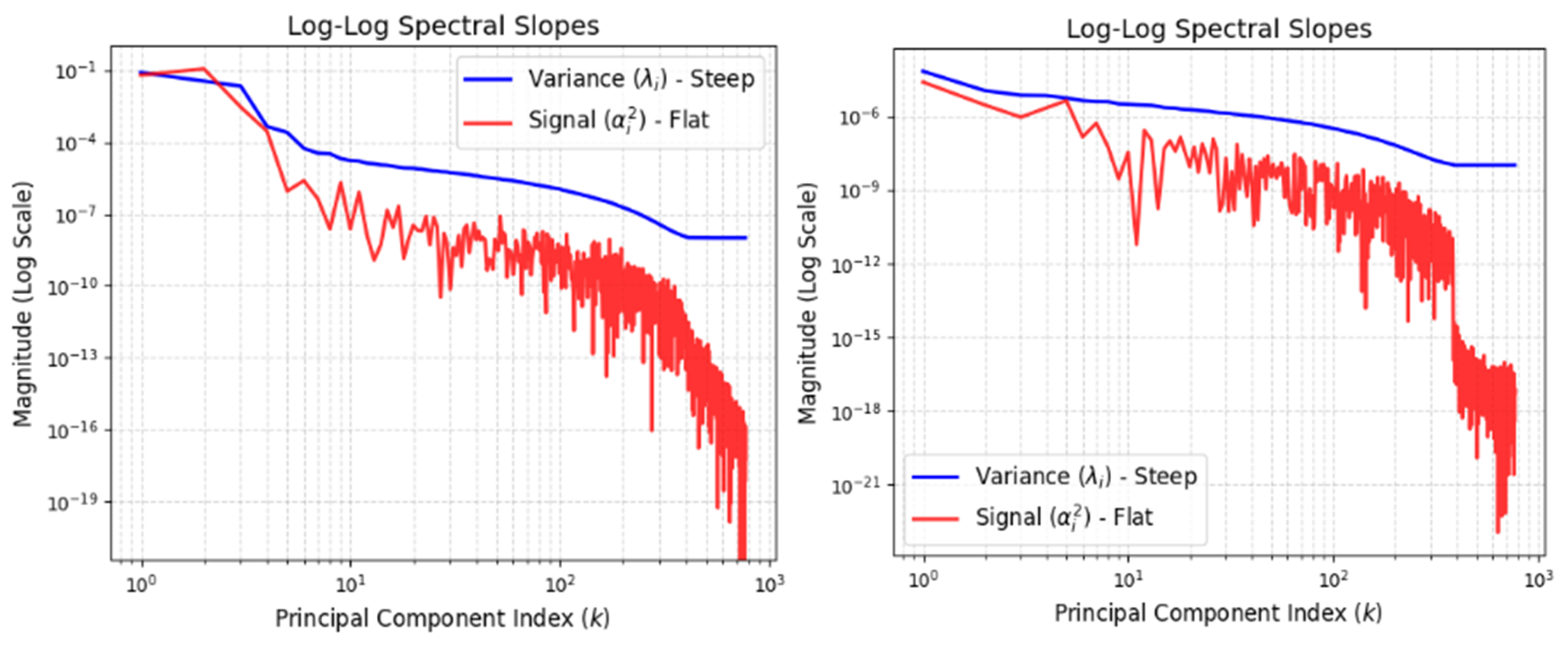}
    \caption{\textbf{Spectral decomposition (log--log) of the embedding covariance for stroke vs.\ rest classification.} VLM (left) and Vision only (right). The blue curve ($\lambda_i$) shows the total variance captured by each mode; the red curve ($\alpha_i^2$) shows the task-relevant signal aligned with the classification objective. Variance decays smoothly across many modes, but the usable signal drops more rapidly, so only a small subset of leading modes contribute to class separation. This gap between variance and signal widens in the spectral tail, where modes carry variance but little discriminative information, illustrating how noise-dominated directions degrade representation quality.}
\end{figure}

By contrast, label-aware spectral metrics isolate the signal-bearing subspace. The structural dimensionality $k(N)$ excludes modes below the recoverability threshold, so it does not track the inflation of the noise bulk. In Table \ref{tab:dcformer_starvation_results}, the Stroke subspace retains a small but nonzero set of stable modes across data regimes, whereas the Healthy Control (CN) subspace collapses from $k(N)=2$ at $100\%$ to $k(N)=0$ at $75\%$ and below. This collapse coincides with a loss of Mahalanobis energy (from $3.22$ to $2.28$), indicating that the class separation is no longer supported by stable directions in the embedding. As $N$ decreases, the signal is  confined to modes within the noise-dominated tail, losing subspace stability and degrading discriminative performance.

To assess if this behavior depends on the specific architecture, we perform an ablation with a convolutional baseline (Table \ref{tab:clip_results_dense}). The CNN exhibits stronger implicit regularization of the covariance spectrum, limiting the inflation of global variance measures such as Effective Rank. However, the underlying signal subspace follows a similar trend: as $N$ decreases, the number of recoverable modes contracts and the Mahalanobis energy decreases (from $0.937$ ROC-AUC at $100\%$ to $0.824$ at $25\%$). The degradation arises from finite-sample spectral effects, the shrinking eigengap and the encroachment of the noise floor, rather than from architectural details. Data efficiency is governed by the stability of the recoverable spectrum, and the transition in performance is a geometric limitation shared across encoder classes.

We illustrate these spectral effects using UMAP projections (see Appendix \ref{sec:appendix_umap}) across data regimes - $100\%$ and $5\%$.  As $N$ decreases, the number of recoverable modes $k(N)$ contracts, the eigengap narrows, and class-discriminative directions fall into the noise-dominated tail. Lower Mahalanobis separation leads to the observed loss of spatial structure. 

\begin{table*}[t]
\centering
\small
\setlength{\tabcolsep}{2pt}
\caption{\textbf{Spectral Dynamics and Predictive Performance in Limited Data Regime (Neuroimaging).}  Comparison of text-anchored VLM vs. unconstrained CNN.}
\label{tab:clip_results_dense}
\begin{tabular}{@{}l ccccccc@{}}
\toprule
\textbf{Metric} & \textbf{5\%} & \textbf{10\%} & \textbf{25\%} & \textbf{35\%} & \textbf{50\%} & \textbf{75\%} & \textbf{100\%} \\
\midrule
\multicolumn{8}{c}{\textbf{Architecture: VLM with CNN as vision encoder}} \\
\midrule
Eff. Rank & 4.500$_{\pm0.500}$ & 4.000$_{\pm2.000}$ & 2.000$_{\pm0.000}$ & 3.000$_{\pm1.000}$ & 3.000$_{\pm0.000}$ & 5.000$_{\pm0.000}$ & 4.500$_{\pm0.500}$ \\
$k(N)$ (Stroke) & 2.000$_{\pm0.000}$ & 1.500$_{\pm1.500}$ & 2.500$_{\pm0.500}$ & 2.000$_{\pm2.000}$ & 4.000$_{\pm0.000}$ & 3.000$_{\pm2.000}$ & 1.000$_{\pm0.000}$ \\
$k(N)$ (AD) & 1.500$_{\pm0.500}$ & 1.500$_{\pm0.500}$ & 1.000$_{\pm0.000}$ & 2.000$_{\pm1.000}$ & 2.000$_{\pm0.000}$ & 5.000$_{\pm0.000}$ & 5.500$_{\pm0.500}$ \\
$k(N)$ (CN) & 0.000$_{\pm0.000}$ & 0.000$_{\pm0.000}$ & 0.000$_{\pm0.000}$ & 1.500$_{\pm1.500}$ & 3.500$_{\pm0.500}$ & 3.000$_{\pm0.000}$ & 0.000$_{\pm0.000}$ \\
M-E (Stroke) & 5.514$_{\pm0.076}$ & 5.859$_{\pm0.131}$ & 5.947$_{\pm0.042}$ & 6.309$_{\pm0.058}$ & 6.332$_{\pm0.035}$ & 6.468$_{\pm0.012}$ & 6.433$_{\pm0.008}$\\
M-E (AD) & 3.780$_{\pm0.088}$ & 4.057$_{\pm0.160}$ & 4.173$_{\pm0.044}$ & 4.316$_{\pm0.012}$ & 4.357$_{\pm0.001}$ & 4.465$_{\pm0.027}$ & 4.436$_{\pm0.014}$ \\
M-E (CN) & 3.064$_{\pm0.064}$ & 3.365$_{\pm0.112}$ & 3.453$_{\pm0.024}$ & 3.653$_{\pm0.033}$ & 3.654$_{\pm0.039}$ & 3.802$_{\pm0.015}$ & 3.743$_{\pm0.015}$ \\
Test AUC & 0.528$_{\pm0.020}$ & 0.594$_{\pm0.049}$ & 0.793$_{\pm0.030}$ & 0.884$_{\pm0.021}$ & 0.903$_{\pm0.002}$ & 0.931$_{\pm0.007}$ & 0.936$_{\pm0.001}$ \\
OOD Test AUC & 0.503$_{\pm0.076}$ & 0.565$_{\pm0.115}$ & 0.688$_{\pm0.005}$ & 0.712$_{\pm0.022}$ & 0.780$_{\pm0.007}$ & 0.770$_{\pm0.001}$ & 0.760$_{\pm0.005}$ \\
\midrule
\multicolumn{8}{c}{\textbf{Architecture: CNN (vision only)}} \\
\midrule
Eff. Rank & 2.500$_{\pm0.500}$ & 3.500$_{\pm0.500}$ & 4.500$_{\pm0.500}$ & 4.500$_{\pm0.500}$ & 3.500$_{\pm0.500}$ & 4.000$_{\pm1.000}$ & 3.500$_{\pm0.500}$ \\
$k(N)$ (Stroke) & 4.500$_{\pm0.500}$ & 3.500$_{\pm0.500}$ & 2.000$_{\pm0.000}$ & 1.500$_{\pm0.500}$ & 1.500$_{\pm0.500}$ & 2.500$_{\pm0.500}$ & 1.000$_{\pm0.000}$ \\
$k(N)$ (AD) & 1.000$_{\pm0.000}$ & 2.000$_{\pm1.000}$ & 2.000$_{\pm0.000}$ & 2.000$_{\pm0.000}$ & 2.000$_{\pm0.000}$ & 2.000$_{\pm1.000}$ & 3.500$_{\pm0.500}$ \\
$k(N)$ (CN) & 0.000$_{\pm0.000}$ & 0.000$_{\pm0.000}$ & 1.500$_{\pm0.500}$ & 2.000$_{\pm0.000}$ & 2.500$_{\pm0.500}$ & 1.500$_{\pm1.500}$ & 3.000$_{\pm1.000}$ \\
M-E (Stroke) & 5.259$_{\pm0.155}$ & 5.762$_{\pm0.239}$ & 5.809$_{\pm0.167}$ & 6.259$_{\pm0.054}$ & 6.339$_{\pm0.016}$ & 6.318$_{\pm0.010}$ & 6.416$_{\pm0.111}$ \\
M-E (AD) & 3.919$_{\pm0.043}$ & 4.085$_{\pm0.057}$ & 3.681$_{\pm0.578}$ & 4.026$_{\pm0.190}$ & 4.085$_{\pm0.078}$ & 4.223$_{\pm0.166}$ & 4.175$_{\pm0.025}$ \\
M-E (CN) & 3.014$_{\pm0.063}$ & 3.221$_{\pm0.148}$ & 2.967$_{\pm0.479}$ & 3.385$_{\pm0.133}$ & 3.459$_{\pm0.056}$ & 3.551$_{\pm0.122}$ & 3.544$_{\pm0.021}$ \\
Test AUC & 0.774$_{\pm0.013}$ & 0.869$_{\pm0.000}$ & 0.815$_{\pm0.058}$ & 0.911$_{\pm0.018}$ & 0.912$_{\pm0.024}$ & 0.934$_{\pm0.025}$ & 0.929$_{\pm0.009}$ \\
OOD Test AUC & 0.565$_{\pm0.003}$ & 0.708$_{\pm0.051}$ & 0.672$_{\pm0.006}$ & 0.761$_{\pm0.035}$ & 0.783$_{\pm0.012}$ & 0.764$_{\pm0.021}$ & 0.773$_{\pm0.021}$ \\
\bottomrule
\end{tabular}
\end{table*}

\section{Discussion}

This work introduces a spectral framework to analyze and design representation learning systems under finite-sample constraints. Performance degradation in low-data regimes is traced to geometry: recoverable spectral structure is lost. As $N$ decreases, the noise floor rises and the set of modes satisfying $\lambda_k \gtrsim \|\hat{\boldsymbol{\Sigma}} - \boldsymbol{\Sigma}\|_{\mathrm{op}}$ contracts, reducing $K(N)$ and destabilizing the learned subspace. This explains why variance-based dimensionality measures can be misleading. The eigengap maintains stability, as classification performance is linked to the distribution of signal across eigenmodes. It separates variance-rich but noise-dominated directions from a smaller set of signal-bearing modes that govern class separation. The framework also supports prediction: estimating spectral decay and $K(N)$ allows us to anticipate future scaling with data, identify crossover points between model classes, and determine if we need additional data or stronger inductive constraints. Multimodal alignment offers spectral regularization, concentrating variance into stable, signal-aligned modes and preserving the eigengap when data is scarce. This motivates architectures and estimators that  control the spectrum, including spectral filtering and zeta-based weighting. The result is a shift from optimizing accuracy alone to controlling representation geometry to boost data efficiency, stability, and interpretability in data-limited domains such as clinical neuroimaging.
\section{Limitations and Future Work}

    Our findings suggest several future directions. A key goal is to formalize the crossover phenomenon by predicting the critical sample size $N_{crit}$ where higher-capacity, spectrally stabilized encoders overtake simpler unimodal models that benefit from inductive bias at very low $N$. Our framework connects to random matrix theory, particularly the Baik–Ben Arous–Péché (BBP) phase transition: as the structural dimensionality $K(N)$ grows, signal eigenvalues may emerge from the Marchenko–Pastur bulk in discrete steps. This can be tested through scaling experiments to determine if such transitions appear as abrupt changes in $AUC(N)$. Extending the theory beyond the IID setting is vital, as clinical data are multi-site and heterogeneous. Modeling site-level variability ($N_{sites}$) may enable spectral debiasing methods that remove site-specific structure while preserving the semantic subspace. Several extensions follow. Our current experiments exclude large-scale vision pretraining to isolate the effect of text-based alignment; domain-specific pretraining should improve ROC-AUC but retain spectral stability. The text encoder is also important. Stronger clinical language models may widen the initial eigengap and delay spectral collapse. The framework can be extended beyond single-modality MRI by anchoring multiple imaging modalities (PET, diffusion MRI) to a shared text representation, enforcing cross-modal stability. While multimodal alignment stabilizes embeddings during training, covariance estimation at inference remains sensitive to finite-sample noise. As a post-hoc calibration method, a \textbf{zeta filter} can replace unstable tail eigenvalues with a structured power-law decay (see Appendix \ref{sec:appendix_zeta}), modestly improving ROC-AUC ($+0.0015$ at $N=2\%$); explicit spectral correction may improve performance without retraining.

\section{Conclusions and Broader Impact}

We introduced a spectral framework for representation learning that links encoder geometry and sample size to downstream performance through the recoverable dimension $K(N)$. Rather than view data efficiency as an empirical property, we show it is governed by the stability of the embedding covariance: as $N$ decreases, the number of modes satisfying $\lambda_k \gtrsim \|\hat{\boldsymbol{\Sigma}}-\boldsymbol{\Sigma}\|_{\mathrm{op}}$ contracts, reducing Mahalanobis separation and degrading classification. $AUC(N)$ is determined by the energy contained in stably recoverable modes. Multimodal alignment offers spectral regularization, concentrating variance into signal-aligned directions, preserving the eigengap, sustaining larger $K(N)$ when data is limited. We unified classical results from random matrix and perturbation theory with modern encoder design, to predict scaling behavior, which may even identify future crossover points between architectures.

Our framework is general. Many computer vision problems involve high-dimensional representations trained with limited supervision, including few-shot learning, domain adaptation, long-tail recognition, and robustness to distribution shift. In each case, performance depends on how signal is distributed across the spectrum and how many modes we can reliably estimate. This suggests that architectures should preserve eigengaps, control tail amplification, and encourage stable low-rank structure. This can be achieved with spectral filtering, zeta-based priors, and covariance-aware training that explicitly controls the embedding spectrum.

In biomedicine, where raw data are high-dimensional (millions of voxels) and sample sizes are typically hundreds to thousands, these issues are  acute. Clinical neuroimaging, genomics, and multimodal phenotyping all suffer from limited sample sizes and heterogeneous data, making spectral instability a  bottleneck. Our framework shows how to boost data efficiency and interpretability by aligning representations to biologically meaningful variation. We will extend these ideas within our ENIGMA Consortium to study how $K(N)$ evolves across diseases, modalities, and populations, and to develop spectrally stable models for diagnosis and biomarker discovery. More broadly, this work suggests a shift from optimizing accuracy to controlling the geometry of learned representations as a primary objective in machine learning.

\section{Acknowledgements}
This work was supported by the U.S. National Institutes of Health, under NIA grant U01 AG068057 and S10OD032285. We are grateful to the ENIGMA Stroke Working Group (https://enigma.ini.usc.edu/ongoing/enigma-stroke-recovery/) for providing the stroke dataset.

\newpage
\appendix

\section{Technical Appendix}

\subsection{Extended Qualitative Results: Subspace Collapse in UMAP Projections}
\label{sec:appendix_umap}

\begin{figure}[htbp]
    \centering
    \includegraphics[width=\textwidth]{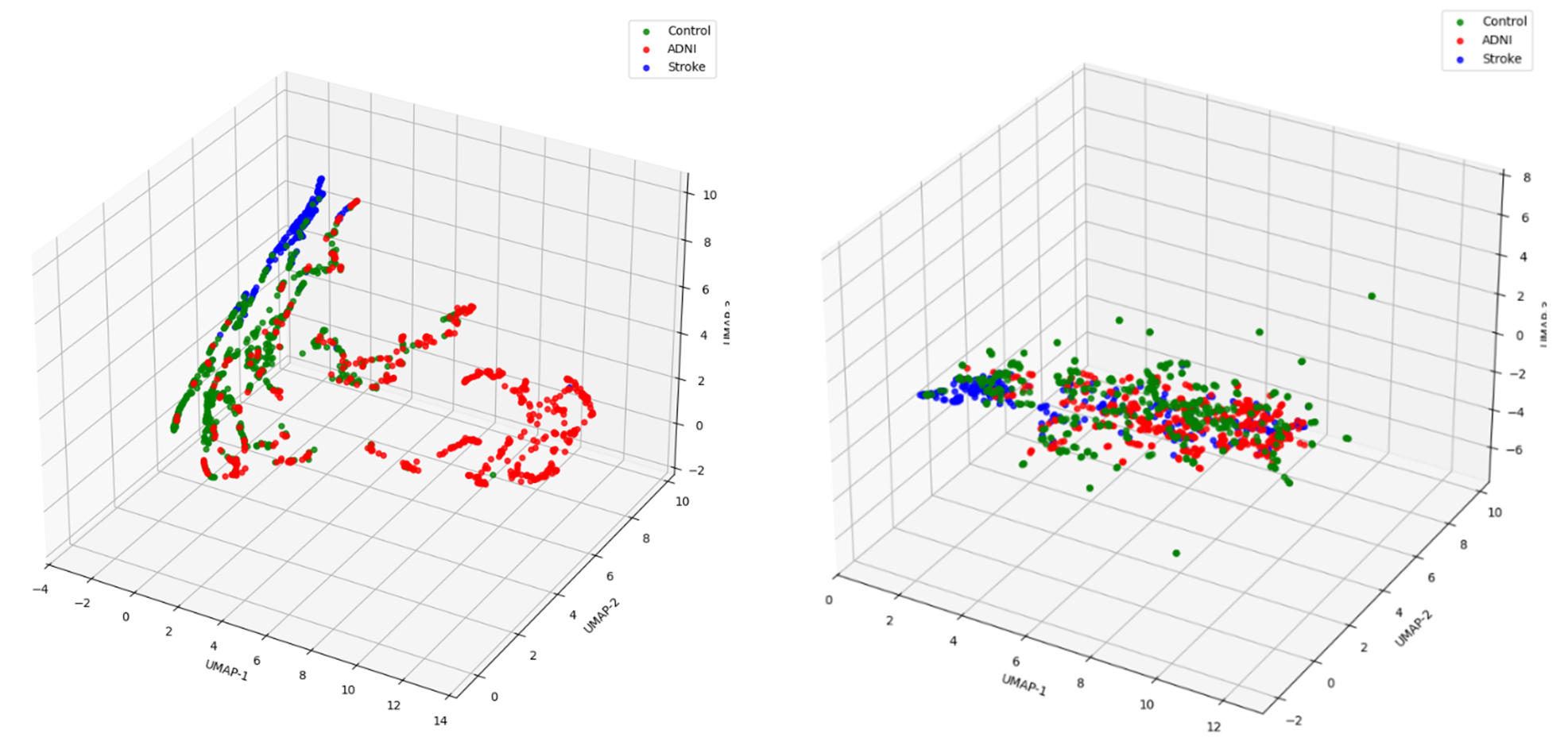}
    \caption{\textbf{UMAP projections of the embedding space across diagnostic classes (VLM + DCFormer).} \textit{Left:} At $N=100\%$, distinct clustering is maintained (Green: Control, Red: Alzheimer's, Blue: Stroke). \textit{Right:} At $N=5\%$, the clustering structure collapses.}
    \label{fig:appendix_umap}
\end{figure}

To visually corroborate the quantitative spectral degradation reported in the main text, we present UMAP projections of the embedding space for the VLM + DCFormer architecture across different data availability regimes (Figure \ref{fig:appendix_umap}). 

At the full $N=100\%$ data regime (Figure \ref{fig:appendix_umap}, Left), the embedding space exhibits distinct, well-separated clusters corresponding to the three diagnostic classes (Healthy Control, Alzheimer's Disease, and Stroke). This spatial separation confirms that the class-discriminative directions are supported by stable, leading eigenmodes of the covariance matrix, well above the finite-sample noise floor. 

Conversely, under severe data starvation at $N=5\%$ (Figure \ref{fig:appendix_umap}, Right), the manifold becomes diffuse and highly overlapping. As our theoretical framework predicts, the contraction of the recoverable dimension $k(N)$ and the narrowing of the eigengap cause the critical class-separating directions to fall into the noise-dominated tail. The qualitative loss of spatial structure in this projection directly mirrors the quantitative decline in Mahalanobis energy and predictive ROC-AUC.

\newpage
\subsection{Post-Hoc Spectral Calibration: The Zeta Filter}
\label{sec:appendix_zeta}

As discussed in the main text, while cross-modal alignment acts as an implicit regularizer during training, empirical covariance estimation during inference remains highly vulnerable to finite-sample noise. To address this without requiring model retraining, we propose a post-hoc spectral calibration technique termed the \textbf{Zeta filter}. 

Concretely, let $\hat{\lambda}_i$ denote the empirical eigenvalues sorted in decreasing order. The zeta filter defines a calibrated spectrum $\tilde{\lambda}_i$ by setting $\tilde{\lambda}_i = \hat{\lambda}_i$ for $i \leq K(N)$ and $\tilde{\lambda}_i = c\, i^{-\beta}$ for $i > K(N)$, where $c$ ensures continuity at $i = K(N)$. 

This differs from classical Wiener filtering, which operates on coefficients and shrinks them according to the signal-to-noise ratio, $\hat{s}_i = \frac{S_i}{S_i + N_i} y_i$. In contrast, the zeta filter acts on the covariance spectrum, replacing the noisy tail with a structured decay before computing Mahalanobis distances or downstream classifiers. Preliminary results on Unconstrained Variance MNIST (Table 5, Figure 6) show modest improvements in ROC-AUC ($+0.0015$ at $N=2\%$), suggesting that explicit spectral correction could potentially improve performance without retraining.

\begin{table}[htbp]
    \centering
    \caption{\textbf{Predictive Performance using Zeta Calibration (Unconstrained Variance MNIST)}.}
    \label{tab:appendix_zeta_results}
    \begin{tabular}{lcccc}
        \toprule
        \textbf{VLM} & \textbf{0.5\%} & \textbf{1\%} & \textbf{2\%} & \textbf{100\%} \\
        \midrule
        ROC-AUC (raw)             & 0.9680 & 0.9827 & 0.9901 & 0.9982 \\
        ROC-AUC ($\zeta$-calibrated) & 0.9632 & 0.9828 & 0.9916 & 0.9987 \\
        \bottomrule
    \end{tabular}
\end{table}

\begin{figure}[htbp]
    \centering
    \includegraphics[width=0.6\linewidth]{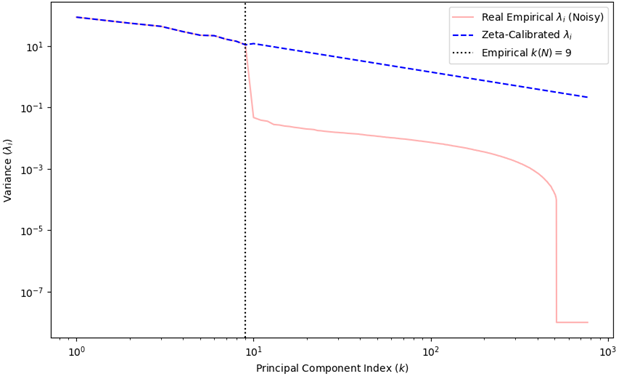}
    \caption{\textbf{Spectral Calibration using the Zeta Filter.} By replacing the noisy empirical tail with a deterministic power-law decay, the filter stabilizes the geometry of the embedding space for downstream performance calculations.}
    \label{fig:appendix_zeta_plot}
\end{figure}

\newpage
\subsection{Robustness to Missing Demographic Modalities}
\label{sec:appendix_masking}

To evaluate the robustness of our multimodal alignment framework to missing clinical data, a common challenge in real-world healthcare systems, we performed ablation studies masking the textual demographic tokens (Age and Sex) using VLM with DCFormer setup. We simulated both a complete modality absence (100\% label drop) and a realistic sparse-data scenario (20\% label drop). As shown in Tables \ref{tab:masking_100} and \ref{tab:masking_20}, the model maintains high discriminative performance across all disease classes even when demographic data is entirely masked. Notably, classification remains highly stable, confirming that our model primarily relies on the robust geometric features extracted from the neuroimaging data rather than simply learning demographic priors.

\begin{table}[h]
    \centering
    \caption{ROC AUC after masking data using VLM - 100\% label drop.}
    \label{tab:masking_100}
    \begin{tabular}{lcccc}
        \toprule
        \textbf{Masked Data} & \textbf{Multiclass} & \textbf{AD vs Rest} & \textbf{Stroke vs Rest} & \textbf{Control vs Rest} \\
        \midrule
        Original (All Labels) & 0.9270 & 0.9260 & 0.9760 & 0.8790 \\
        Age and Sex        & 0.9095 & 0.9206 & 0.9783 & 0.8296 \\
        Age                   & 0.9140 & 0.9180 & 0.9783 & 0.8434 \\
        Sex                & 0.9358 & 0.9394 & 0.9716 & 0.8963 \\
        \bottomrule
    \end{tabular}
\end{table}

\begin{table}[h]
    \centering
    \caption{ROC AUC after masking data using VLM - 20\% label drop.}
    \label{tab:masking_20}
    \begin{tabular}{lcccc}
        \toprule
        \textbf{Masked Data} & \textbf{Multiclass} & \textbf{AD vs Rest} & \textbf{Stroke vs Rest} & \textbf{Control vs Rest} \\
        \midrule
        Original (All Labels) & 0.9270 & 0.9260 & 0.9760 & 0.8790 \\
        Age and Sex        & 0.9193 & 0.9217 & 0.9740 & 0.8621 \\
        Age                   & 0.9188 & 0.9194 & 0.9746 & 0.8623 \\
        Sex               & 0.9265 & 0.9282 & 0.9728 & 0.8785 \\
        \bottomrule
    \end{tabular}
\end{table}

\clearpage
\section{Supplementary Material}
\paragraph{B1. AUC–Mahalanobis Relation (Gaussian Case)}

We first link classification performance to geometry. Assume class-conditional features $\mathbf{x} \in \mathbb{R}^D$ follow Gaussians with shared covariance:
\[
\mathbf{x}\mid y=0 \sim \mathcal{N}(\boldsymbol{\mu}_0, \boldsymbol{\Sigma}), \quad 
\mathbf{x}\mid y=1 \sim \mathcal{N}(\boldsymbol{\mu}_1, \boldsymbol{\Sigma}).
\]
The optimal linear classifier projects onto $\mathbf{w} = \boldsymbol{\Sigma}^{-1}(\boldsymbol{\mu}_1 - \boldsymbol{\mu}_0)$. The resulting signal-to-noise ratio is
\[
d_M^2 = (\boldsymbol{\mu}_1 - \boldsymbol{\mu}_0)^T \boldsymbol{\Sigma}^{-1} (\boldsymbol{\mu}_1 - \boldsymbol{\mu}_0),
\]
the Mahalanobis distance. Under Gaussian assumptions, the AUC is
\[
\mathrm{AUC} = \Phi\!\left(\frac{d_M}{2}\right),
\]
so errors in estimating $\boldsymbol{\Sigma}$ directly translate into errors in AUC.

\paragraph{B2. Spectral Decomposition of Mahalanobis Distance}
\label{sec:spec_decomp_maha}

Let $\boldsymbol{\Sigma} = V \Lambda V^T$ with eigenvalues $\lambda_i$ and eigenvectors $\mathbf{v}_i$. Writing $\mathbf{d} = \boldsymbol{\mu}_1 - \boldsymbol{\mu}_0$ and $\alpha_i = \mathbf{v}_i^T \mathbf{d}$,
\[
d_M^2 = \sum_{i=1}^{D} \frac{\alpha_i^2}{\lambda_i}.
\]
This decomposition separates signal ($\alpha_i^2$) from variance ($\lambda_i$). Modes with small $\lambda_i$ amplify noise, so the tail of the spectrum can dominate $d_M^2$ even when it carries little signal.

\paragraph{B3. Finite-Sample Perturbation (Davis--Kahan)}

In practice, $\boldsymbol{\Sigma}$ is replaced by $\hat{\boldsymbol{\Sigma}}$. The Davis--Kahan theorem bounds the error in eigenvectors:
\[
\sin \Theta(\mathbf{v}_k, \hat{\mathbf{v}}_k) 
\le 
\frac{2 \|\hat{\boldsymbol{\Sigma}} - \boldsymbol{\Sigma}\|_{\mathrm{op}}}{\delta_k},
\]
where $\delta_k$ is the eigengap. As $\delta_k$ decreases, eigenvectors become unstable and lose alignment with the true signal directions.

\paragraph{B4. Recoverability of Spectral Modes}

Combining the spectral decomposition with perturbation theory yields a condition for recoverability. A mode $k$ remains stable only if its eigenvalue dominates the perturbation:
\[
\lambda_k \gtrsim \|\hat{\boldsymbol{\Sigma}} - \boldsymbol{\Sigma}\|_{\mathrm{op}}.
\]
Modes that violate this condition undergo eigenvector mixing and no longer represent the true geometry. This defines the effective cutoff $K(N)$ of recoverable modes.

\paragraph{B5. Scaling of the Noise Floor (Vershynin)}

For sub-Gaussian data, concentration bounds give
\[
\|\hat{\boldsymbol{\Sigma}} - \boldsymbol{\Sigma}\|_{\mathrm{op}} 
= \mathcal{O}\!\left(\sqrt{\frac{D}{N}}\right).
\]
Substituting into the recoverability condition,
\[
\lambda_k \gtrsim \sqrt{\frac{D}{N}}.
\]
As $N$ decreases, the noise floor rises and $K(N)$ contracts, pushing more modes into the noise-dominated regime. This explains spectral collapse and motivates tail regularization (e.g., Zeta-filtering), which replaces unstable empirical eigenvalues in the unresolved region $\lambda_k < \sqrt{D/N}$ with a structured decay.

The preceding results can be summarized as a single finite-sample spectral law linking covariance estimation, recoverable dimension, and classification performance:

\begin{center}
\fbox{
\begin{minipage}{0.92\linewidth}

\colorbox{black}{\parbox{\linewidth}{\color{white}\textbf{Zeta Law: From Covariance Estimation to Recoverable Modes}}}

\vspace{6pt}

Let $\boldsymbol{\Sigma}=V\Lambda V^\top$ be the population embedding covariance, with eigenvalues $\lambda_i$ and class-difference projections $\alpha_i=\mathbf{v}_i^\top(\boldsymbol{\mu}_1-\boldsymbol{\mu}_0)$. The Mahalanobis energy decomposes across spectral modes as
\[
d_M^2=\sum_{i=1}^{D}\frac{\alpha_i^2}{\lambda_i}.
\]

With finite samples, the empirical covariance $\hat{\boldsymbol{\Sigma}}$ perturbs the spectrum. For sub-Gaussian embeddings, the Vershynin concentration formula gives the noise scale
\[
\|\hat{\boldsymbol{\Sigma}}-\boldsymbol{\Sigma}\|_{\mathrm{op}}
\sim \sqrt{\frac{D}{N}}.
\]

The Davis--Kahan theorem then implies that mode $k$ is stable only when its eigenvalue exceeds this perturbation scale:
\[
\lambda_k \gtrsim \|\hat{\boldsymbol{\Sigma}}-\boldsymbol{\Sigma}\|_{\mathrm{op}}.
\]
This defines the recoverable dimension $K(N)$, the number of modes that can be reliably used at sample size $N$. The usable Mahalanobis energy is truncated to the stable modes:
\[
d_M^2(N)\approx \sum_{i=1}^{K(N)}\frac{\alpha_i^2}{\lambda_i}.
\]
If the signal-aligned spectral energy follows a power law, $\alpha_i^2/\lambda_i \sim i^{-\beta}$, then
\[
d_M^2(N)\sim \sum_{i=1}^{K(N)}i^{-\beta}
\;\to\; \zeta(\beta)
\quad \text{as } K(N)\to\infty,
\]
where $\zeta$ denotes the Riemann zeta function.

\end{minipage}
}
\end{center}

\newpage
\section{Experimental Details and Hyperparameters}
\label{sec:appendix_hps}

To ensure full reproducibility of the spectral analysis and classification results presented in the main text, we detail the hyperparameters used for both the cross-modal VLM and the unconstrained vision baselines (Table \ref{tab:hyperparameters}). All models were trained using PyTorch on NVIDIA V100 GPUs. Training for the VLM was performed using DeepSpeed ZeRO-2 to enable memory-efficient distributed optimization.

\begin{table}[htbp]
    \centering
    \caption{\textbf{Training Hyperparameters.} Default configurations used across the Unconstrained Variance MNIST and Neuroimaging experiments unless otherwise specified.}
    \label{tab:hyperparameters}
    \small
    \begin{tabular}{ll}
        \toprule
        \textbf{Hyperparameter} & \textbf{Value} \\
        \midrule
        Optimizer & AdamW \\
        Learning Rate (Vision Encoder) & $1 \times 10^{-4}$ \\
        Learning Rate (Text/Projection) & $1 \times 10^{-4}$ \\
        Weight Decay (Vision Encoder) & $0.001$ \\
        Weight Decay (Text/Projection) & $0.1$ \\
        Batch Size & 16 \\
        Training Epochs & 20 \\
        Learning Rate Schedule & Cosine Annealing with Linear Warmup \\
        Warmup Epochs (Vision Encoder) & 1 \\
        Warmup Epochs (Text/Projection) & 0.6 \\
        Loss Function & Pairwise Sigmoid Loss / Cross-Entropy \\
        \bottomrule
    \end{tabular}
\end{table}

\end{document}